\documentclass[letterpaper]{article} % DO NOT CHANGE THIS
\usepackage{aaai23}  % DO NOT CHANGE THIS
\usepackage{times}  % DO NOT CHANGE THIS
\usepackage{helvet}  % DO NOT CHANGE THIS
\usepackage{courier}  % DO NOT CHANGE THIS
\usepackage[hyphens]{url}  % DO NOT CHANGE THIS
\usepackage{graphicx} % DO NOT CHANGE THIS
\usepackage{natbib}  % DO NOT CHANGE THIS AND DO NOT ADD ANY OPTIONS TO IT
\usepackage{caption} % DO NOT CHANGE THIS AND DO NOT ADD ANY OPTIONS TO IT
\usepackage{algorithm}
\usepackage{algorithmic}

\usepackage{amsmath} %对齐公式
\usepackage{subcaption} %子图包
\usepackage{threeparttable} %三线栏包，其实可以画成任何一行你想要画线分割的行处
\usepackage{multirow} %跨行跨列必用包
\usepackage{booktabs} %漂亮表的包
\usepackage{bbding} %对错符号包
\usepackage{color}
\usepackage{amssymb}

\usepackage{newfloat}
\usepackage{listings}
\DeclareCaptionStyle{ruled}{labelfont=normalfont,labelsep=colon,strut=off} % DO NOT CHANGE THIS
\floatstyle{ruled}
\newfloat{listing}{tb}{lst}{}
\floatname{listing}{Listing}
\title{Resolving Task Confusion in Dynamic Expansion Architectures for \\ Class Incremental Learning}
\author{
    Bingchen Huang\textsuperscript{\rm 1,\rm 2},
    Zhineng Chen\textsuperscript{\rm 1,\rm 2}\thanks{Zhineng Chen is the corresponding author.},
    Peng Zhou\textsuperscript{\rm 3},
    Jiayin Chen\textsuperscript{\rm 1,\rm 2},
    Zuxuan Wu\textsuperscript{\rm 1,\rm 2}
}
\affiliations{
    \textsuperscript{\rm 1}Shanghai Key Lab of Intelligent Information Processing, School of Computer Science, Fudan University\\
    \textsuperscript{\rm 2}Shanghai Collaborative Innovation Center on Intelligent Visual Computing\\
    \textsuperscript{\rm 3}University of Maryland, College Park, MD, USA\\
    \{bchuang21, jiayinchen21\}@m.fudan.edu.cn, \{zhinchen, zxwu\}@fudan.edu.cn, pengzhou@terpmail.umd.edu
}

\usepackage{bibentry}
\begin{document}

\maketitle

\begin{abstract}
The dynamic expansion architecture is becoming popular in class incremental learning, mainly due to its advantages in alleviating \emph{catastrophic forgetting}. However, task confusion is not well assessed within this framework, e.g., the discrepancy between classes of different tasks is not well learned (i.e., inter-task confusion, ITC), and certain priority is still given to the latest class batch (i.e., old-new confusion, ONC). We empirically validate the side effects of the two types of confusion. Meanwhile, a novel solution called \emph{Task Correlated Incremental Learning} (TCIL) is proposed to encourage discriminative and fair feature utilization across tasks. TCIL performs a multi-level knowledge distillation to propagate knowledge learned from old tasks to the new one. It establishes information flow paths at both feature and logit levels, enabling the learning to be aware of old classes. Besides, attention mechanism and classifier re-scoring are applied to generate more fair classification scores. We conduct extensive experiments on CIFAR100 and ImageNet100 datasets. The results demonstrate that TCIL consistently achieves state-of-the-art accuracy. It mitigates both ITC and ONC, while showing advantages in battle with catastrophic forgetting even no rehearsal memory is reserved. \footnote{Source code: https://github.com/YellowPancake/TCIL}
\end{abstract}

\section{Introduction}
Class incremental learning aims to develop machine learning algorithms that are capable of continuously learning new classes, while retaining the knowledge learned from old classes~\cite{m32,t54}. It receives increasing attention as the learning is in line with the underlying assumption in many real-world applications, e.g., the classes to be processed dynamically evolve through time~\cite{d25}, previous data are unavailable for privacy-preserving reasons~\cite{d19, e0}. Recently, deep neural networks have been applied to this field and have achieved impressive performance. However, these studies commonly suffer from the problem of \emph{catastrophic forgetting}, i.e., the optimization of model parameters caused by new task learning, meanwhile, often leads to decreased performance on previously learned old classes~\cite{m12,t38,lu2022bridging}.

Many studies are proposed to fight with catastrophic forgetting. They include: constraining weight changes~\cite{t38, e1, p24, e2, w20}, retaining an extra memory that stores a certain amount of previous data~\cite{s43, s14, w5, v29, e2}, synthesizing training data~\cite{e4,e5,zhu2021prototype,zhu2022self}, etc. Recently, the dynamic expansion architecture (DEA) is emerging as a promising paradigm~\cite{t23, t27, t33, t70, t47, t0}. It dynamically expands the network as the number of tasks increases, where each task is associated with a dedicated sub-network and its weights are frozen when learning a new task. It has the advantage of keeping the knowledge of old tasks well. Currently, leaderboards of public benchmarks are dominated by DEA-based models. 

\begin{figure}
\centering
    \begin{subfigure}[t]{0.48\textwidth}
          \centering
            \includegraphics[width=\linewidth]{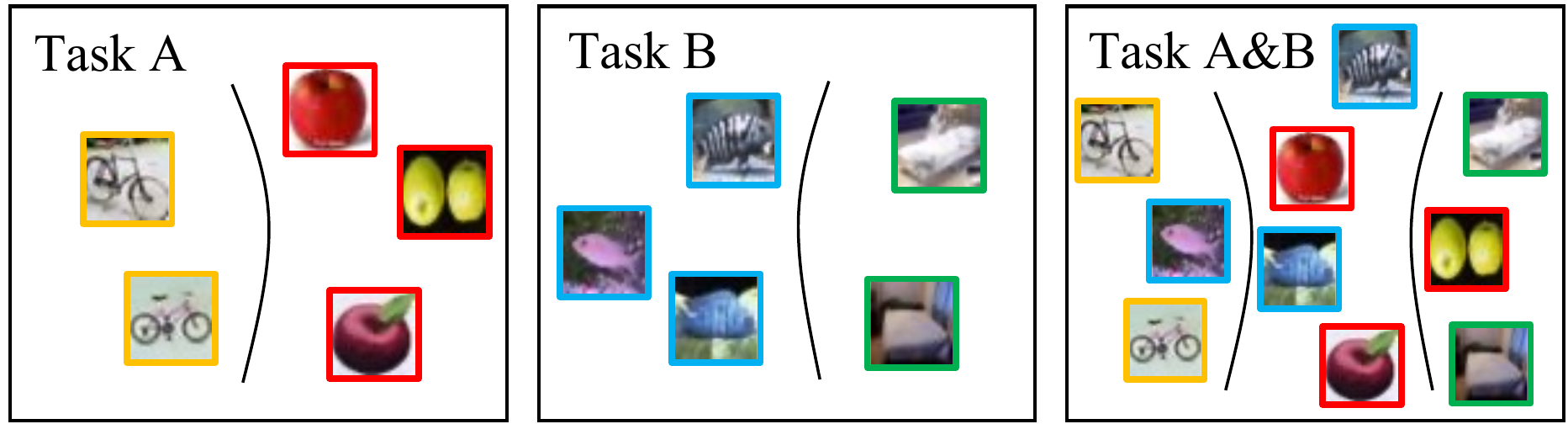}
            \caption{inter-task confusion (ITC)}
            \label{img1_a}
    \end{subfigure}
    \begin{subfigure}[t]{0.25\textwidth}
            \centering
            \includegraphics[width=\textwidth]{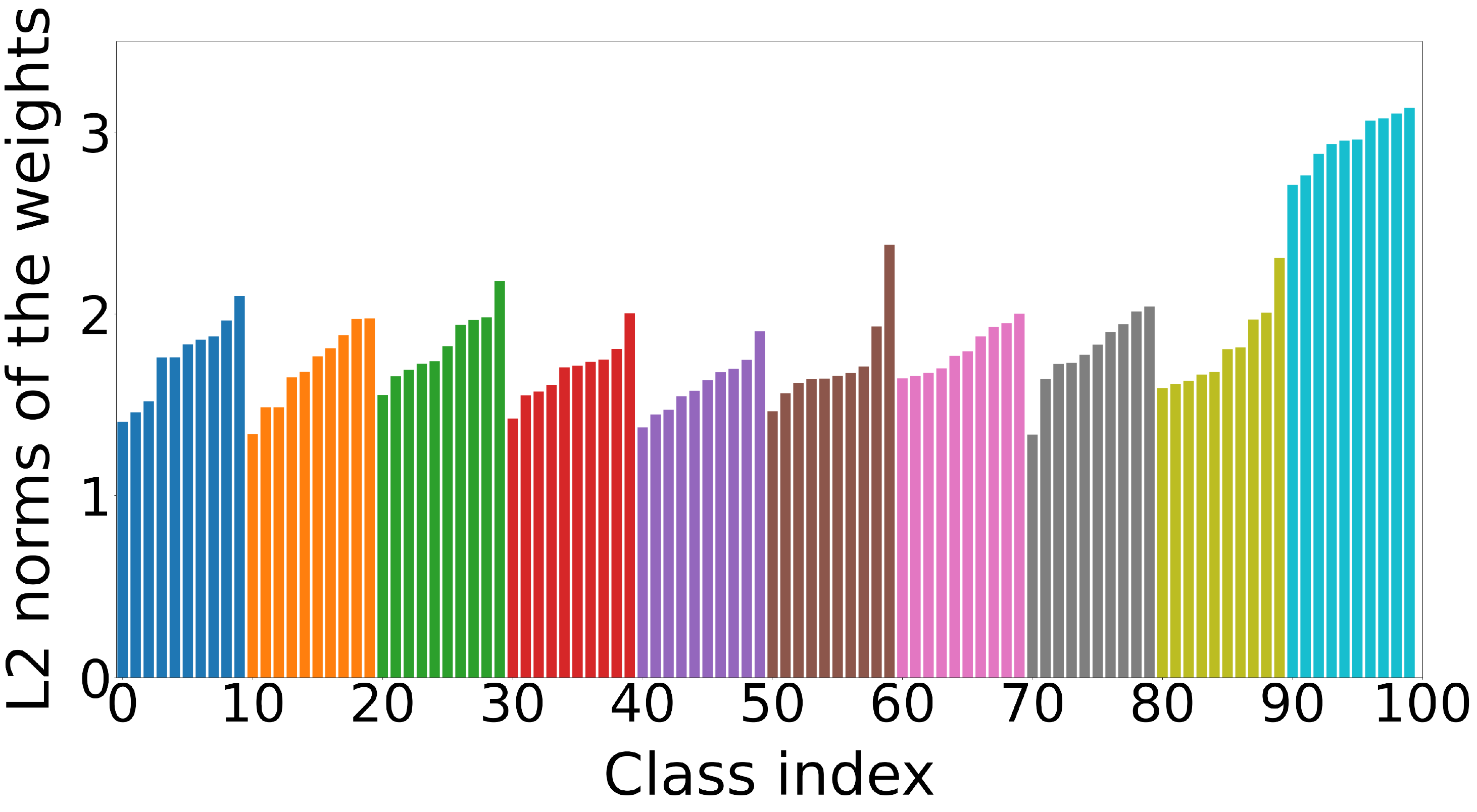}
            \caption{old-new confusion (ONC)}
            \label{img1_b}
    \end{subfigure}
    \begin{subfigure}[t]{0.20\textwidth}
            \centering
            \includegraphics[width=\textwidth]{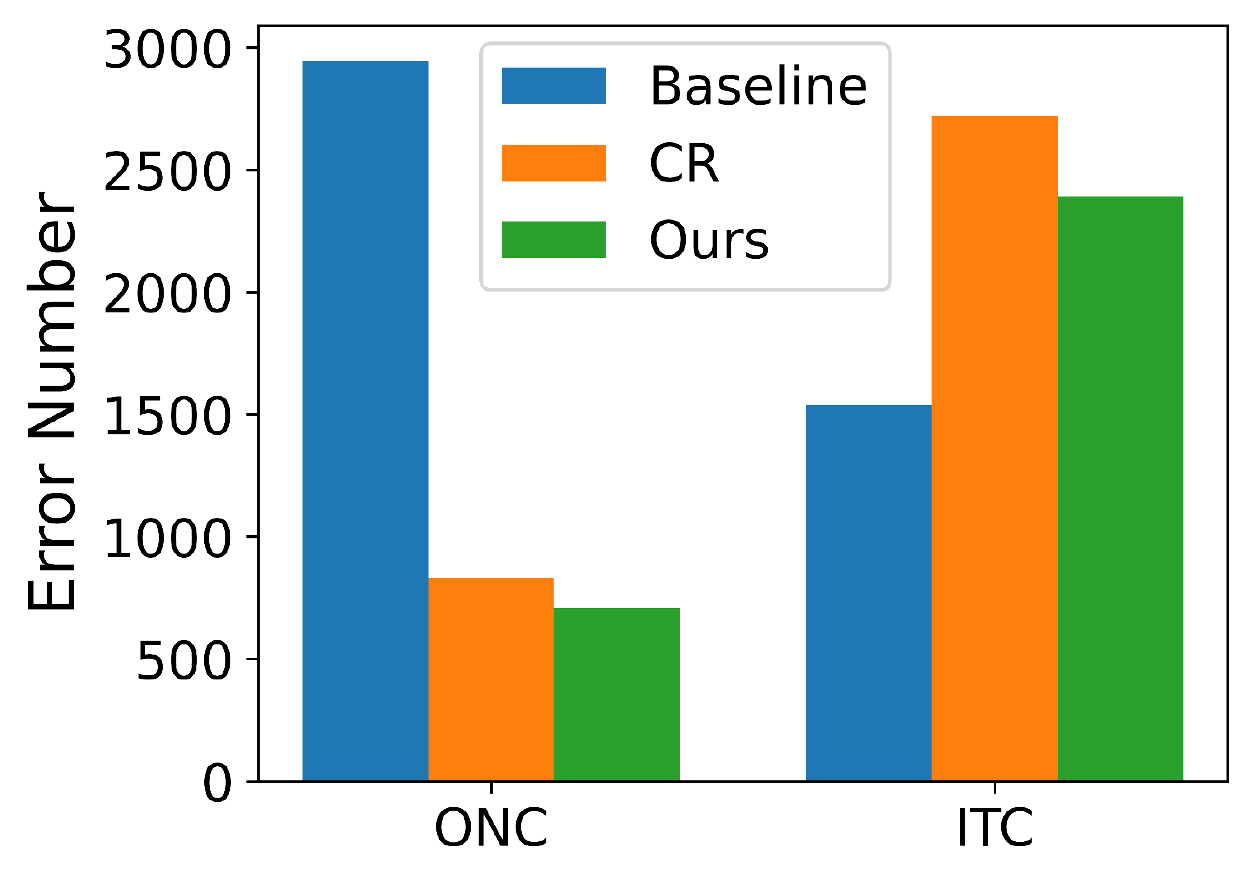}
            \caption{error statistics}
            \label{img1_c}
    \end{subfigure}
\caption{Task confusion illustration. (a) ITC: class discrepancy within every task is well learned, but it is not taught to distinguish classes from different tasks. (b) ONC: old classes have similar weight distribution while larger weights are given to new classes. (c) TCIL significantly reduces ONC errors on CIFAR100. It also suppresses the cases that ONC transforming to ITC to some extent (CR V.S. Ours), leading to remarkable overall confusion reduction. }
\end{figure}

However, DEAs are unable to process two kinds of confusion effectively due to their dynamically expandable nature. The first is \emph{inter-task confusion} (ITC). An illustrative example is shown in Fig.\ref{img1_a}. Both tasks \emph{A} and \emph{B} are trained to distinguish their own classes. However, the discrepancy between classes from different tasks is not taught when the two tasks are combined, thus causing confusion. The second is \emph{old-new confusion} (ONC). It is explained as the classifier would give priority to new classes rather than old classes. Since tasks share the same classification layer whose weights are lastly optimized by new classes, the weights are oftentimes dominated by new classes, as shown in Fig.\ref{img1_b}. We argue that both ITC and ONC caused by the correlation between tasks are barely investigated in DEAs. 

Note that the correlation between tasks in class incremental learning has been explored in the literature~\cite{v0, t69, t32}, for example, in the form of knowledge distillation~\cite{w20, t54, t19, min2020multi}. However, knowledge distillation does not always lead to improvements~\cite{v0, v25}, and it is not trivial to establish due to the characteristic of DEA, which makes the correlation between tasks in DEAs less explored. Moreover, while ONC has been widely recognized as the task-tendency bias~\cite{v25, t69, d39}, existing DEAs address it mainly by finetuning the classifier on a balanced training subset, which brings additional training burden and is highly dependent on the memory budget. Simple but effective solutions such as weight aligning \cite{d39} have not yet been carefully investigated in DEAs. Besides, a majority of existing dynamic expansion methods are with certain prerequisites, e.g., requiring a task identifier at testing~\cite{t23, t33, t67}, sensitive to rehearsal memory~\cite{t70}, or needing complex hyperparameter tuning~\cite{t70}. 

% Note that the correlation between tasks in incremental learning has been explored in literature~\cite{v0, t69, t32}. For example, in the form of knowledge distillation~\cite{w20, t54, t19}. However, knowledge distillation does not always lead to improvements~\cite{v0, v25}. Moreover, the distillation is not trivial to establish as the characteristic of DEA. On the other hand, while ONC has been widely recognized as the task-tendency bias~\cite{v25, t69, d39}, existing DEAs address it mainly by leveraging the rehearsal memory. Simple but effective solutions such as weight aligning \cite{d39} has not yet been carefully investigated. Besides, a majority of existing dynamic expansion methods are with certain prerequisites, e.g., requiring a task identifier at testing~\cite{t23, t33, t67}, sensitive to rehearsal memory~\cite{t70}, or needing complex hyperparameter tuning~\cite{t70}. 

Motivated by the aforementioned issues, we propose a novel framework termed as \emph{Task Correlated Incremental Learning} (TCIL) that aims to mitigate catastrophic forgetting from the angle of resolving task confusion in the DEA framework. As shown in Fig.\ref{img1_c}, ONC is still the major error type in CIFAR100. To this end, a classifier re-scoring (CR) strategy is applied to rectify the heavily biased classification layer. Similar to~\cite{d39}, it calculates the weight magnitude statistics according to old and new classes. Moreover, in contrast to previous DEAs that directly concatenate old and new features together, we design a feature fusion module to attend to the most relevant features by using the attention mechanism. With these modifications, ONC errors are largely reduced but it also causes an increase in ITC errors. Therefore, a novel multi-level knowledge distillation is developed to further deal with ITC, where knowledge propagation mechanisms are established at both feature and logit levels. Specifically, it builds information propagation paths from every old feature extractor to the new extractor, generating distillations from previous classifiers to the current classifier. It forms a complete supervision from old knowledge to the new classifier. As seen in Fig.\ref{img1_c}, it reduces the errors in both ITC and ONC (CR V.S. Ours). By combining all the upgrades, TCIL is capable of better handling task confusion and thus catastrophic forgetting, producing a more discriminative and fair classification. To validate the effectiveness of TCIL, we conduct extensive experiments on CIFAR100 and ImageNet100 datasets with variants such as different task splits, with or without rehearsal memory, etc. The results demonstrate that TCIL consistently achieves state-of-the-art accuracy. While TCIL-Lite, the lite version TCIL, is smaller but still effective. Moreover, compared with existing methods, TCIL is more robust as the increase of incremental steps and is less sensitive to the availability of rehearsal memory, e.g., showing greater accuracy gaps in case no rehearsal memory is kept.

Contributions of this paper can be summarized as follows. We analyze the classification error and task confusion within the DEA framework. It mainly consists of ITC and ONC. To this end, we propose TCIL, a novel scheme to promote a discriminative and fair feature utilization across tasks in class incremental Learning. Specifically, we integrate a classifier re-scoring strategy along with a feature fusion module to alleviate ONC. Meanwhile, a multi-level knowledge distillation is developed to further suppress ITC. As a result, TCIL greatly mitigates the two types of task confusion, it consistently performs top-tier and shows advantages in solving catastrophic forgetting even in non-rehearsal setting.

%To avoid ONC and integrate with dynamic extension network, we adapt a classifier re-scoring strategy used in static architectures and a feature fusion module.

\begin{figure*}
\centering
   \includegraphics[width=0.85\linewidth,height=0.28\textheight]{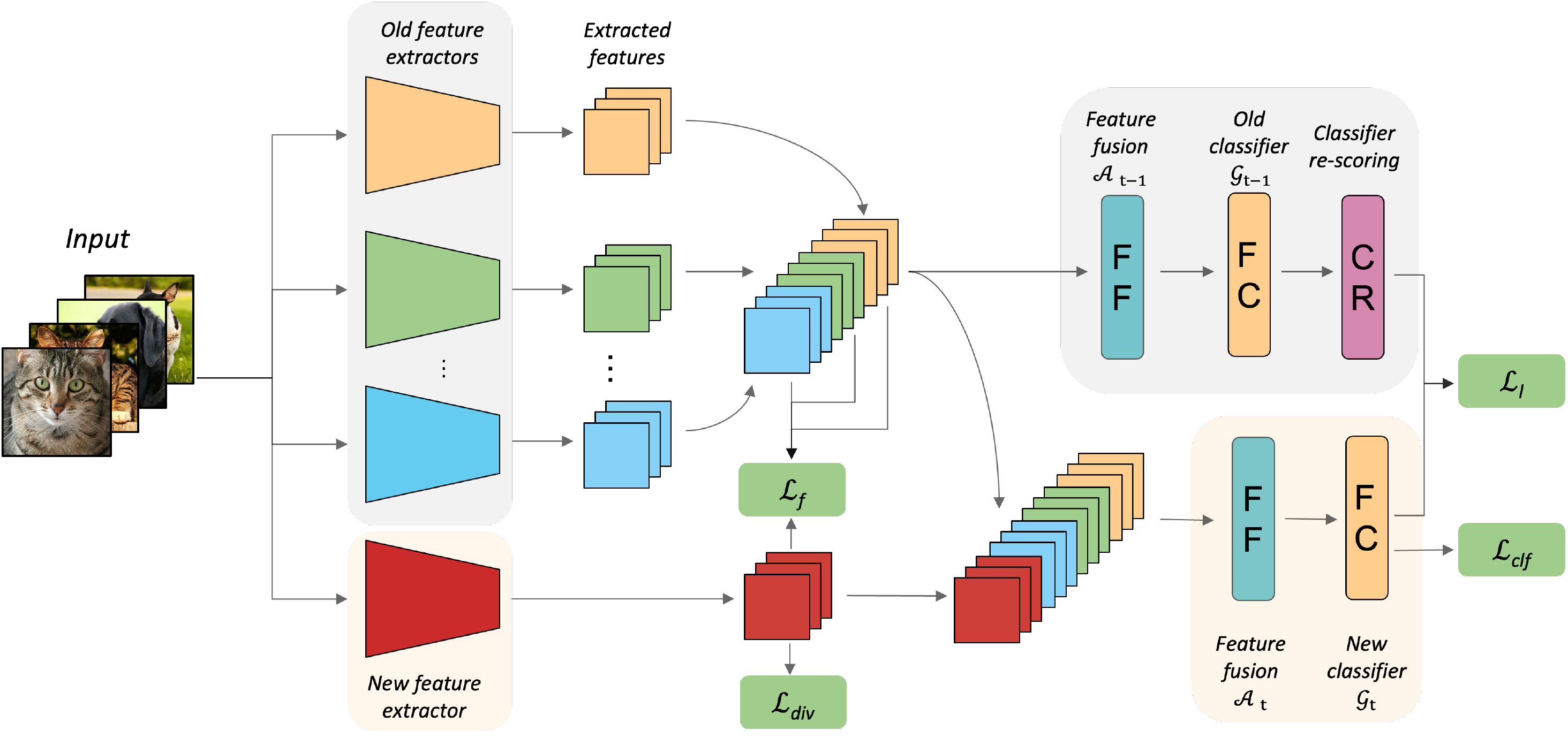}

%   \caption{An illustrative framework of the proposed TCIL. It uses a dedicated feature extraction sub-network for each specific task. $\mathcal{L}_{\mathrm{f}}$ and $\mathcal{L}_{\mathrm{l}}$ are knowledge distillation loss at feature and logit levels, respectively. $\mathcal{L}_{\mathrm{clf}}$ is the classification loss.}
    \caption{An illustrative framework of the proposed TCIL. It uses a dedicated feature extraction sub-network for each specific task. $\mathcal{L}_{\mathrm{f}}$ and $\mathcal{L}_{\mathrm{l}}$ are knowledge distillation loss at feature and logit levels, respectively. $\mathcal{L}_{\mathrm{clf}}$ is the classification loss, and $\mathcal{L}_{\mathrm{div}}$ is the divergence loss for guiding the training of the feature extractors.}
   \label{img1}
\end{figure*}

\section{Related Work}

\textbf{Catastrophic Forgetting.}
Many researches have been carried out to overcome catastrophic forgetting. We can broadly classify them into three categories: regularization-based~\cite{t38, e1, p24, w20}, rehearsal-based~\cite{s43, s14, w5, v29, e2} and DEA-based~\cite{t23, t27, t33, t70, t47, t0}. Regularization-based methods emphasize on constraining the weight changes. For example, allowing only small magnitude changes on previous weights. It suffers from the problem that the changes cannot sufficiently describe the complex pattern shift caused by new task learning. Rehearsal-based methods reserve a small amount of old data when training a new task. By doing so, it retains certain previous knowledge. Studies in this category mainly focus on the selection of old data and the way to use it. For example, iCaRL was developed to learn an exemplar-based data representation~\cite{t54}. However, rehearsal methods are difficult to scale up to a large number of classes because of the memory constraint, e.g., only a limited number of training samples could be reserved in total. Nevertheless, strategies such as generating synthetic data~\cite{e3,e5} or features~\cite{e4} also alleviated this dilemma. 
%To ease the classification, rehearsal with limited data memory still holds for a majority of incremental learning methods.  

Alternatively, DEAs choose a different way that dynamically creates feature extraction sub-networks each associated with one specific task~\cite{t23, t27, t33, t56, t11, t67}. Early methods required a task identifier to select the right subset of parameters at test-time. Unfortunately, the assumption is unrealistic as new samples would not come with their task identifiers. Recently, DER~\cite{t70} proposed a dynamically expandable representation by discarding the task identifier, where the classifier was finetuned on a balanced exemplar subset to mitigate the task-tendency bias. Li~\cite{t47} also proposed a multi-extractor based learning framework, where knowledge distillation and network pruning were leveraged. However, the distillation was applied to nearby tasks only. DyTox~\cite{t0} shared encoder and decoder among tasks while differentiating tasks only by special tokens. It largely reduced the network size and attained impressive results.

\noindent\textbf{Knowledge Distillation.}
It aims to utilize a large teacher model to guide the training of a small student model~\cite{r9,yang2022multi,yang2022rd}. Performing the learning at logit-level is an effective and direct way of knowledge distillation~\cite{e14}. It is expected to distillate previous learned old knowledge to new task model in class incremental learning. LwF~\cite{w20} first applied it to this scenario, where a modified cross-entropy loss was used to preserve the capabilities of old model. It was extended to multi-class classification scenarios later~\cite{t54}. M2KD~\cite{w39} introduced a multi-level knowledge distillation strategy. It utilized all previous model snapshots instead of distilling knowledge only from the last model. However, it is investigated under fixed network structure. Recently, the distillation was considered from intermediate layers rather than the outputted logit, by keeping either feature map activation~\cite{p24}, the spatial pooling~\cite{t19}, or the normalized global pooling~\cite{t32} as similar as possible. However, these methods were not well combined with the solution for task-tendency bias yet, thus restricting the classification accuracy to some extent. 

Note that DEAs are different from previous incremental learning paradigms. Both the task correlation and task-tendency bias are not well investigated in the DEA framework. We formulate the issues as ITC and ONC and propose to use multi-level knowledge distillation and classifier re-scoring to address them.

\section{Method}

\subsection{Problem Formulation and Method Overview}
%Our goal is to learn a unified model that will classify an increasingly growing number of classes. 
We first describe the problem investigated in the context of image classification. Assume there are $T$ batches of data $\left\{\mathcal{D}_{1}, \cdots, \mathcal{D}_{T}\right\}$, with $\mathcal{D}_{t}=\left\{\left(\mathbf{x}_{1}^{t}, y_{1}^{t}\right), \cdots,\left(\mathbf{x}_{n_{t}}^{t}, y_{n_{t}}^{t}\right)\right\}$ as the training data at step $t$ (i.e., task $t$), where ${x}_{i}^{t}$ is the $i$-th input image and $y_{i}^{t} \in \mathcal{C}_{t}$ is the label within the label set $\mathcal{C}_{t}$, $n_{t}$ is the number of samples in set $\mathcal{D}_{t}$. At the $t$-th incremental step, training batches $t$ will be added to the training set. Therefore, the goal can be formulated as to learn knowledge from new data $\mathcal{D}_{t}$, while retain the previous knowledge learned from old data $\left\{\mathcal{D}_{1}, \cdots, \mathcal{D}_{t-1}\right\}$. The label space of the model is all seen categories $\tilde{\mathcal{C}}_{t}=\cup_{i=1}^{t} \mathcal{C}_{i}$ and the model is expected to predict well on all classes in $\tilde{\mathcal{C}}_{t}$. All label sets are exclusive, i.e., $\mathcal{C}_{i} \cap \mathcal{C}_{j}=\emptyset$ for arbitrarily given $i$ and $j$. In rehearsal setting, an exemplar subset of the previous data under fixed memory budget is accessible for every incremental steps. While access to previous data is forbidden in non-rehearsal setting.

To mitigate catastrophic forgetting, we propose TCIL based on the DEA framework. Its overview is illustrated in Fig.\ref{img1}. As seen, a multi-level knowledge distillation is established at both feature and logit levels, i.e., $\mathcal{L}_{\mathrm{f}}$ and $\mathcal{L}_{\mathrm{l}}$. It encourages information propagation from every old task to the new task, thus alleviating ITC and ONC. Meanwhile, feature extracted from different tasks are appropriately fused to highlight these important ones. While weights of the outputted layer are adjusted to derive a more fair classification, thus mitigating ONC. The pipeline is elaborated as follows.

\subsection{TCIL Architecture and Details}

\textbf{TCIL Pipeline.} 
In DEAs, the feature extractor $\mathcal{F}_{1}$ and classifier $\mathcal{G}_{1}$ at the first step is trained the same as previous methods~\cite{p24, e2, w20, t54}. Then at each step $t \in \left\{2, \cdots, T\right\}$, we add a new feature extractor $\mathcal{F}_{t}$ while keeping the parameters of previous extractors $\left\{\mathcal{F}_{1}, \cdots, \mathcal{F}_{t-1}\right\}$ and previous classifier $\mathcal{G}_{t-1}$ frozen. Meanwhile, we initialize the parameters of $\mathcal{G}_{t}$ with $\mathcal{G}_{t-1}$. Given an image $x$ from the seen batches $\left\{1, \cdots, t\right\}$, we concatenate all extracted features $\boldsymbol{u}_{t}$ as follows:
\begin{equation}
\boldsymbol{u}_{t}
=\left[\mathcal{F}_{1}(\boldsymbol{x}), \cdots, \mathcal{F}_{t}(\boldsymbol{x})\right]
\label{e1}
\end{equation}
The new feature extractor $\mathcal{F}_{t}$ learns from both ${D}_{t}$, the $t$-th data batch and $\boldsymbol{u}_{t-1}$, the feature representation at step $t-1$, by using the proposed feature-level knowledge distillation elaborated later. Then we get the refined features $\boldsymbol{f}_{t}$  by an attention-based feature fusion module $\mathcal{A}_t$ (also elaborated later) as follows:
\begin{equation}
\boldsymbol{f}_{t}=\mathcal{A}_t(\boldsymbol{u}_{t}) 
\label{e2}
\end{equation}
Then we feed the refined features $\boldsymbol{f}_{t}$ into the new classifier $\mathcal{G}_{t}$, and get the output logits $\boldsymbol{o}_{t}(x)$. During model training, a logit-level knowledge distillation is also applied to guide $\mathcal{G}_{t}$ preserving old knowledge.
\begin{equation}
\boldsymbol{o}_{t}(\boldsymbol{x})=\mathcal{G}_{t}(\boldsymbol{f}_{t})
\label{e3}
\end{equation}
While at inference, a classifier re-scoring module in the form of $\mathcal{W}_{t}$ for the new task is figured out at the end of each training step, namely:
\begin{equation}
\boldsymbol{o}_{t}(\boldsymbol{x})=\mathcal{W}_{t}\left(\mathcal{G}_{t}(\boldsymbol{f}_{t})\right)
\label{e32}
\end{equation}
The new model learns probability distribution from the previous step, and we can get the the prediction $\hat{y}$ for image $x$ to calculate the cross-entropy loss:
\begin{equation}
\begin{aligned}
p_{\mathcal{G}_{t}}(\boldsymbol{y} \mid \boldsymbol{x})&=\operatorname{Softmax}\left(\boldsymbol{o}_{t}(\boldsymbol{x})\right) \\
\hat{y}&=\arg \max p_{\mathcal{G}_{t}}(\boldsymbol{y} \mid \boldsymbol{x})
\label{e4}
\end{aligned}
\end{equation}

\noindent\textbf{Multi-level Knowledge Distillation (MLKD).}
%We apply knowledge distillation to both feature and classifier to preserving old knowledge learned from previous steps.
In rehearsal setting, let $\mathcal{R}$ be the set of reserved samples. We apply the feature-level knowledge distillation to assist the learn of $\mathcal{F}_{t}$, the new feature extractor. Specifically, given a reserved image $\boldsymbol{x}\in \mathcal{D}_{i}$, we use the $i$-th feature extractor $\mathcal{F}_{i}$ as the teacher to guide the training, the feature-level knowledge distillation loss can be represented as:

\begin{equation}
\mathcal{L}_{f}(\boldsymbol{x})=||\boldsymbol{F}_{t}(\boldsymbol{x})- \boldsymbol{F}_{i}(\boldsymbol{x})||_2, 
\label{e6}
\end{equation}

Then we explain how the logit-level knowledge distillation is formulated. For each training image  $\boldsymbol{x}$ we can get $\boldsymbol{o}_{t}(\boldsymbol{x})$ and $\boldsymbol{o}_{t-1}(\boldsymbol{x})$, the outputs of classifiers $\mathcal{G}_{t}$ and $\mathcal{G}_{t-1}$, respectively. Then, we use KL divergence~\cite{r23} to calculate the distance between them:

\begin{equation}
\mathcal{L}_{l}(\boldsymbol{x})=\sum_{c=1}^{\tilde{\mathcal{C}}_{t-1}}{q}_{c}(\boldsymbol{x}) \log \left(\frac{{q}_{c}(\boldsymbol{x})}{\hat{q}_{c}(\boldsymbol{x})}\right),
\label{e7}
\end{equation}
where $\hat{q}_{c}(\boldsymbol{x})=\frac{e^{\hat{o}_{c}(x) / T}}{\sum_{j=1}^{\tilde{\mathcal{C}}_{t-1}} e^{\hat{o}_{j}(x) / T}}$, $ q_{c}(\boldsymbol{x})=\frac{e^{o_{c}(x) / T}}{\sum_{j=1}^{\tilde{\mathcal{C}}_{t-1}} e^{o_{j}(x) / T}}$, $T$ is the temperature scalar. $\hat{o}_{c}(x)$ is an element of $\boldsymbol{o}_{t-1}(\boldsymbol{x})$, $\boldsymbol{o}_{t-1}(\boldsymbol{x})=(\hat{o}_{1}(\boldsymbol{x}), \cdots, \hat{o}_{\tilde{\mathcal{C}}_{t-1}}(\boldsymbol{x}))^{T}$ represents the logits of $\mathcal{G}_{t-1}$. $o_{c}(x)$ is an element of $\boldsymbol{o}_{t}(\boldsymbol{x})$, $\boldsymbol{o}_{t}(\boldsymbol{x}) = (o_{1}(\boldsymbol{x}), \cdots, o_{\tilde{\mathcal{C}}_{t-1}}(\boldsymbol{x}), o_{\tilde{\mathcal{C}}_{t-1}+1}(\boldsymbol{x}), \cdots, o_{\tilde{\mathcal{C}}_{t}}(\boldsymbol{x}))^{T}$ stands for the logits of $\mathcal{G}_{t}$. Then, parameters of both feature extractor $\mathcal{F}_{t}$ and classifier $\mathcal{G}_{t}$ are updated with the combined loss during training. The total loss can be written as:
\begin{equation}
\mathcal{L}_{\mathrm{kd}}= \lambda \sum_{\boldsymbol{x}\in \mathcal{R}} \mathcal{L}_{f}(\boldsymbol{x}) + \mu  \sum_{\boldsymbol{x}\in \mathcal{R}\cup\mathcal{D}_{t}} \mathcal{L}_{l}(\boldsymbol{x}),
\label{e88}
\end{equation}
where $\lambda$ and $\mu$ are hyperparameters both empirically set to 0.5. $\lambda=0$ means the non-rehearsal setting, i.e., all loss is from $\mathcal{L}_{l}$.

\begin{table*}
\centering
\resizebox{\linewidth}{!}{
\begin{tabular}{@{}l|ccc|ccc|ccc|ccc|ccc|ccc@{}}    
\toprule
\multicolumn{1}{c|}{\multirow{3}{*}{\textbf{Methods}}} &
\multicolumn{9}{c|}{CIFAR100-B0} &
\multicolumn{9}{c}{CIFAR100-B50} \\ \cmidrule(l){2-19} 

% \multicolumn{1}{c|}{\multirow{2}{*}{\textbf{Methods}}}
\multicolumn{1}{c|}{} 
& \multicolumn{3}{c|}{5 steps}                                     & \multicolumn{3}{c|}{10 steps}                                    & \multicolumn{3}{c|}{20 steps}                                     &
\multicolumn{3}{c|}{2 steps}                                     & \multicolumn{3}{c|}{5 steps}                                     & \multicolumn{3}{c}{10 steps}                                     \\ \cmidrule(l){2-19} 
\multicolumn{1}{c|}{}                                  & \textbf{\#Paras} & \textbf{Avg}          & \textbf{Last}         & \textbf{\#Paras} & \textbf{Avg}          & \textbf{Last}         & \textbf{\#Paras} & \textbf{Avg}          & \textbf{Last}         &
\textbf{\#Paras} & \textbf{Avg}          & \textbf{Last}         & \textbf{\#Paras} & \textbf{Avg}          & \textbf{Last}         & \textbf{\#Paras} & \textbf{Avg}          & \textbf{Last}         
\\ \midrule
Bound                                                  & 11.2             & 80.40                 & -                     & 11.2             & 80.41                 & -                     & 11.2             & 81.49                 & -     & 11.2             & 77.20                  & -                     & 11.2             & 79.89                 & -                     & 11.2             & 79.91                 & -                    \\ \midrule
iCaRL~\shortcite{t54}                                                  & 11.2             & 71.14                 & 59.71                 & 11.2             & 65.27                 & 50.74                 & 11.2             & 61.20                 & 43.75                 & 11.2             & 77.20                  & -                     & 11.2             & 79.89                 & -                     & 11.2             & 79.91                 & -                    \\
UCIR~\shortcite{t32}                                                   & 11.2             & 62.77                 & 47.31                 & 11.2             & 58.66                 & 43.39                 & 11.2             & 58.17                 & 40.63                & 11.2             & 67.21                 & 56.82                 & 11.2             & 64.28                 & 52.02                 & 11.2             & 59.92                 & 48.02                 \\
BiC~\shortcite{t69}                                                    & 11.2             & 73.10                 & 62.1                  & 11.2             & 68.80                 & 53.54                 & 11.2             & 66.48                 & 47.02                & 11.2             & 72.47                 & 64.22                 & 11.2             & 66.62                 & 55.01                 & 11.2             & 60.25                 & 48.04                \\
RPSNet~\shortcite{t52}                                                 & 60.6             & 70.50                 & 60.56                 & 56.5             & 68.60                 & 57.05                 & -                & -                     & -                     & -                & -                     & - & -                & -                     & - & -                & -                     & -\\
WA~\shortcite{d39}                                                     & 11.2             & 72.81                 & 60.84                 & 11.2             & 69.46                 & 53.78                 & 11.2             & 67.33                 & 47.31                & 11.2             & 71.43                 & 62.37                 & 11.2             & 64.01                 & 52.87                 & 11.2             & 57.86                 & 47.90                \\
PODNet~\shortcite{t19}                                                 & 11.2             & 66.70                 & 51.71                 & 11.2             & 58.03                 & 41.05                 & 11.2             & 53.97                 & 35.02                & 11.2             & 71.30                  & 62.11                 & 11.2             & 67.25                 & 55.94                 & 11.2             & 64.04                 & 52.13               \\
DER~\shortcite{t70}                                                    & 33.6             & 76.80                 & 68.32                 & 61.6             & 75.36                 & 65.22                 & 117.6            & 74.09                 & 62.48                & 22.4             & 74.61                 & 68.84                 & 39.2             & 73.21                 & 65.77                 & 67.2             & 72.81                 & 65.45                 \\
DyTox~\shortcite{t0}                                                  & -                & -                     & -                     & 10.7             & 73.66                 & 60.67                 & 10.7             & 72.27                 & 56.32               & -                & -                     & -  & -                & -                     & -  & -                & -                     & -    \\ \midrule
TCIL                                                   & 34.3             & \textbf{77.72} & \textbf{69.58} & 64.1             & \textbf{77.30} & 66.41 & 127.1            & 75.11 & 63.54 & 22.7             & \textbf{76.42} & \textbf{71.91} & 40.2             & \textbf{74.88} & 68.58 & 70.3             & \textbf{73.72} & 66.36\\
TCIL-Lite                                                 & 8.3              & 76.96 & 69.44 & 14.5             & 76.74 & \textbf{66.66} & 28.1             & \textbf{75.47} & \textbf{64.08} & 5.4              & 74.95 & 70.72 & 8.3              & 74.30 & \textbf{68.89} & 14.5             & 73.50 & \textbf{67.26}\\
\midrule

\end{tabular}
}
\caption{Top-1 accuracy comparison on CIFAR100 in rehearsal setting. Dytox~\cite{t0} and RPSNet \cite{t52} results come from their respective papers, and other results come from ~\cite{t70}.}
\label{Cifar-B0}
\end{table*}

\noindent\textbf{Classifier Re-scoring (CR).}\label{CM}
When the step $t > 1$, ONC appears as $\mathcal{G}_{t}$ is biased to new classes. Similar to~\cite{d39}, we propose to re-score the old and new classes based on their weight norms. To be specific, at the end of each training step, we calculate the weight norms of old classes and new classes in the last fully connected layer as follows:

\begin{equation}
\begin{aligned}
 \boldsymbol{n}_{old}&=\left(\left\|\mathbf{w}_{1}\right\|, \cdots,\left\|\mathbf{w}_{\tilde{\mathcal{C}}_{t-1}}\right\|\right),\\
\boldsymbol{n}_{new}&=\left(\left\|\mathbf{w}_{\tilde{\mathcal{C}}_{t-1}+1}\right\|, \cdots,\left\|\mathbf{w}_{\tilde{\mathcal{C}}_{t}}\right\|\right) \label{e8}
\end{aligned}
\end{equation}
Based on the above norms, we calculate the coefficient $\gamma$ for classifier re-scoring:
\begin{equation}
\gamma={\operatorname{Mean}\left(\boldsymbol{n}_{{old }}\right)}/{\operatorname{Mean}\left(\boldsymbol{n}_{{new }}\right)}
\label{e9}
\end{equation}
where $\operatorname{Mean(\cdot)}$ returns the mean value of elements in the vector. We rewrite the output logits $\boldsymbol{o}_{rt}(\boldsymbol{x})$ in the following form: $\boldsymbol{o}(\boldsymbol{x})=(\boldsymbol{o}_{t-1}(\boldsymbol{x}), \boldsymbol{o}_{t}(\boldsymbol{x})[\tilde{\mathcal{C}}_{t-1}+1,\dots,\tilde{\mathcal{C}}_{t}])$, where $\boldsymbol{o}_{t}(\boldsymbol{x})$ indicates the logits of $\mathcal{G}_{t}$. Then the rectified logits $\boldsymbol{o}(\boldsymbol{x})$ is given by:
\begin{equation}
\boldsymbol{o}_{rt}(\boldsymbol{x})=\mathcal{W}_{t}(\boldsymbol{o}(\boldsymbol{x})) = \left(\mathbf{o}_{o l d}(\boldsymbol{x}), \gamma \cdot \boldsymbol{o}_{n e w}(\boldsymbol{x})\right)
\label{e11}
\end{equation}
By doing this, the average norm of outputted logits for new classes becomes the same as those of the old classes, thus well mitigating ONC. Note that within new classes (or old classes), the relative magnitude of logits does not change. 

\noindent\textbf{Feature Fusion Module (FFM).}\label{FFM}
Considering that feature extractors $\left\{\mathcal{F}_{1}, \cdots, \mathcal{F}_{t-1}\right\}$ are trained at different steps and they could not extract features of images from new steps $\left\{t, \cdots, \mathcal{T}\right\}$ well, which affects the quality of features, we apply FFM to refine the features and combine the transformed features for classification. Given an intermediate feature map $\boldsymbol{u}_{t} \in \mathbb{R}^{C \times H \times W}$ as input, similar to ~\cite{c0}, FFM sequentially infers a 1D attention map $\mathbf{A}_{\mathbf{c}} \in \mathbb{R}^{C \times 1 \times 1}$ and a 2D attention map $\mathbf{A}_{\mathbf{s}} \in \mathbb{R}^{1 \times H \times W}$. The whole feature fusion process can be summarized as:
\begin{equation}
\begin{aligned}
\boldsymbol{f}_{t}=\mathcal{A}_t(\boldsymbol{u}_{t}) &=\mathbf{A}_{\mathbf{s}}\left(\mathbf{A}_{\mathbf{c}}(\boldsymbol{u}_{t}) \otimes \boldsymbol{u}_{t}\right) \otimes \left(\mathbf{A}_{\mathbf{c}}(\boldsymbol{u}_{t}) \otimes \boldsymbol{u}_{t}\right)
\end{aligned} \label{e12}
\end{equation}
where $\otimes$ denotes element-wise multiplication. Therefore, $\mathbf{A}_{\mathbf{c}}$ and $\mathbf{A}_{\mathbf{s}}$ are the channel attention~\cite{c28, c0, e10} and spatial attention~\cite{c0, e11} defined as follows.
\begin{equation}
\begin{aligned}
\mathbf{A}_{\mathbf{c}}(\boldsymbol{u}_{t}) &=\sigma(M L P(\operatorname{Avg} \operatorname{Pool}(\boldsymbol{u}_{t}))+M L P(\operatorname{Max} \operatorname{Pool}(\boldsymbol{u}_{t})))
\end{aligned} \label{e13}
\end{equation}
\begin{equation}
\begin{aligned}
\mathbf{A}_{\mathbf{s}}(\boldsymbol{u}_{t}) &=\sigma\left(f^{7 \times 7}([\operatorname{Avg} \operatorname{Pool}(\boldsymbol{u}_{t}) ; \operatorname{Max} \operatorname{Pool}(\boldsymbol{u}_{t})])\right) 
\end{aligned} \label{e14}
\end{equation}
where $\sigma$ denotes the sigmoid function and $f^{7 \times 7}$ represents a convolution with kernel size of $7 \times 7$. During multiplication, the attention values are broadcasted accordingly.

\noindent\textbf{Training Loss.}
TCIL is trained with three losses, i.e., the classification loss $\mathcal{L}_{clf}$ calculated by cross-entropy, the multi-level knowledge distillation loss given by Eq.\ref{e88}, and a divergence loss $\mathcal{L}_{\mathrm{div}}$ to maximum the discrepancy between old-new classes as in~\cite{t70}. Formally, the total loss is:
\begin{equation}
\mathcal{L}= \mathcal{L}_{\mathrm{clf}}+\alpha \mathcal{L}_{\mathrm{kd}}+\beta \mathcal{L}_{\mathrm{div}},
\label{e5}
\end{equation}
where $\alpha$ and $\beta$ are hyperparameters both empirically set to 1.0 for all experiments. 

Note that the proposed multi-level knowledge distillation significantly enriches the use of previous old knowledge. It not only establishes information propagation paths from every old feature extractor to the new extractor at the feature level, but also enables the information sharing from the old classifier to the new one at the logit level, thus forming a multi-grained distillation that has not been proposed before. In addition, the re-scoring strategy is also applied for the first time to the DEA framework.
%It is also different from the "multi-level" knowledge distillation in~\cite{w39} and in ~\cite{t19}, which is not targeted for the DEA framework.

%\noindent\textbf{Network Pruning}
\subsection{Network Pruning}
Since feature extraction sub-networks are sequentially added with new tasks, TCIL parameters grow with the incremental steps, which is undesired in real-world applications. As a solution, we devise TCIL-Lite, the lite version TCIL, by applying parameter pruning techniques described in~\cite{e12}. Instead of pruning small convolution kernels, it calculates the geometric median~\cite{e15} between the kernels. Then kernel pairs whose similarity above a threshold are treated as redundant and one of them is pruned. With this strategy, the model size could be largely reduced while not affecting the accuracy too much. Since pruning is not the emphasis of this paper, we empirically set TCIL-Lite nearly four times smaller than TCIL. We will demonstrate the effectiveness of TCIL-Lite in the experiments.

\begin{figure*}
\centering

\includegraphics[width=1\linewidth]{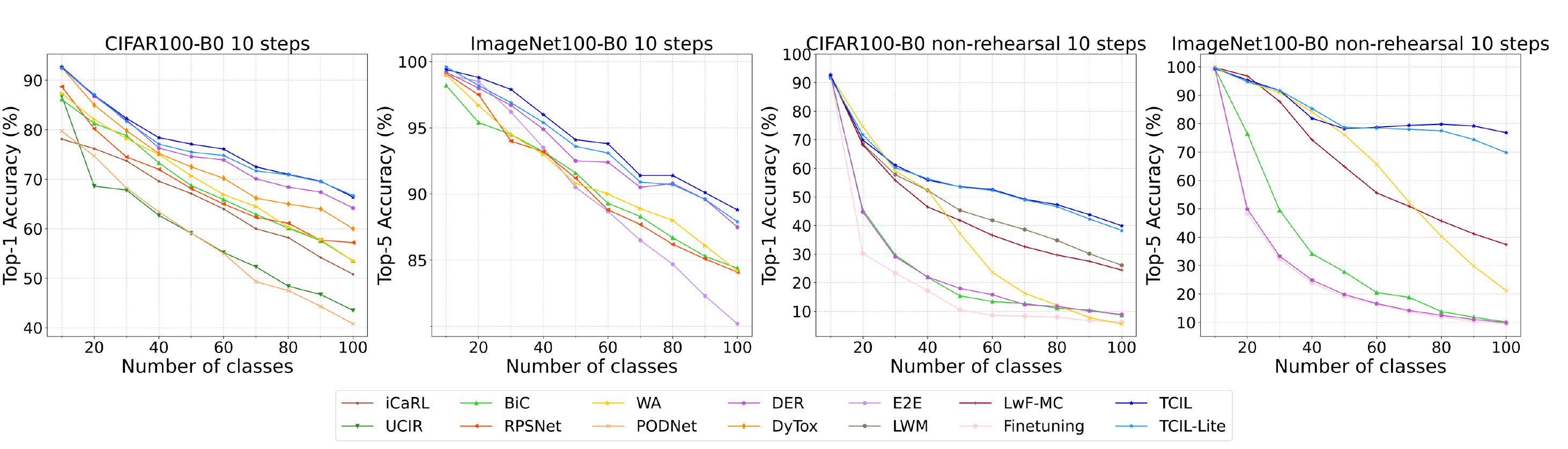}

\caption{Accuracy evolution on benchmarks. The left are evaluated on CIFAR100-B0 and ImageNet100-B0 with 10 steps in rehearsal setting. The right are evaluated on CIFAR100-B0 and ImageNet100-B0 with 10 steps in non-rehearsal setting.}
\label{img2}
\end{figure*}

\section{Experiments}\label{experiment}

\subsection{Implementation}
%To evaluate TCIL, we compare our approach with other methods in both rehearsal and non-rehearsal settings.
To evaluate TCIL, we conduct extensive experiments on CIFAR100~\cite{e6} and ImageNet100~\cite{e7} under two memory settings: fixed exemplar memory (rehearsal) and none exemplar memory (non-rehearsal). In rehearsal setting, following~\cite{t0, t70, t54, t19}, we set the memory size to 2000 images. For CIFAR100, two protocols are considered. The first is CIFAR100-B0: training all 100 classes from scratch with different task splits, i.e., 5, 10 and 20 incremental steps. The second is CIFAR100-B50: starting from a model trained on 50 classes, and the remaining 50 classes are divided into splits of 2, 5 and 10 steps. We report the top-1 accuracy both averaged over the incremental steps (Avg) and after the last step (Last). For ImageNet100, we assess our methods on the ImageNet100-B0 protocol: the model is trained in batches of 10 classes from scratch and uses the same ImageNet subset and class order following~\cite{t19, t32, t54, t70}. Both top-1 and top-5 accuracy is reported. While in non-rehearsal setting, we evaluate TCIL on both CIFAR100-B0 and ImageNet100-B0 protocols. 

Following~\cite{t47, t70}, we adopt ResNet-18 as the basic network for feature extraction. For CIFAR100, we employ random crop and horizontal flip as the data augmentation. While for ImageNet100, we employ the data augmentation in \cite{e13}. Data augmentation like rotation, brightness variation and cutout, are randomly performed during the training. We adopt SGD optimizer with weight decay 0.0005 and batch size 128 for all experiments. We use the warmup strategy with the ending learning rate 0.1 for 10 epochs in CIFAR100 and 20 epochs in ImageNet100, respectively. After the warmup, for CIFAR100 the learning rate is 0.1 and decays to 0.01 and 0.001 at 100 and 120 epochs. For ImageNet100 the learning rate decays to 0.01, 0.001 and 0.0001 at 60, 120 and 180 epochs. In rehearsal setting, we use the same exemplar selection strategy as~\cite{t54, t32}. All models are trained by using a workstation with 1 Nvidia 3090 GPU on Pytorch.

\begin{table}
\centering
\resizebox{0.75\linewidth}{!}{
\begin{tabular}{@{}l|ccccc@{}}
\toprule
\multicolumn{1}{c|}{}                                   & \multicolumn{5}{c}{ImageNet100-B0}                                                                                      \\ \cmidrule(l){2-6} 
\multicolumn{1}{c|}{}                                   & \multicolumn{3}{c|}{\textbf{top-1}}                                     & \multicolumn{2}{c}{\textbf{top-5}}                     \\ \cmidrule(l){2-6} 
\multicolumn{1}{c|}{\multirow{-3}{*}{\textbf{Methods}}} & \textbf{\#Paras} & \textbf{Avg}   & \multicolumn{1}{c|}{\textbf{Last}}  & \textbf{Avg}   & \textbf{Last}                         \\ \midrule
Bound                                                   & 11.2             & -              & \multicolumn{1}{c|}{-}              & 95.10           & -                                     \\ \midrule
iCaRL~\shortcite{t54}                                                   & 11.2             & -              & \multicolumn{1}{c|}{-}              & 83.60           & 63.80                                  \\
E2E~\shortcite{w5}                                                     & 11.2             & -              & \multicolumn{1}{c|}{-}              & 89.92          & 80.29                                 \\
BiC~\shortcite{t69}                                                     & 11.2             & -              & \multicolumn{1}{c|}{-}              & 90.60           & 84.40                                  \\
RPSNet~\shortcite{t52}                                                  & -                & -              & \multicolumn{1}{c|}{-}              & 87.90           & 74.00                                    \\
WA~\shortcite{d39}                                                      & 11.2             & -              & \multicolumn{1}{c|}{-}              & 91.00             & 84.10                                  \\
DER~\shortcite{t70}                                                     & 61.6             & 77.18          & \multicolumn{1}{c|}{66.70}           & 93.23          & 87.52
\\
DyTox~\shortcite{t0}                                                     & 11.0             & 77.15          & \multicolumn{1}{c|}{\textbf{69.10}}           & 92.04          & 87.98 \\ \midrule
TCIL                                                    & 64.1             & \textbf{77.66} & \multicolumn{1}{c|}{67.34} & \textbf{94.17} & \textbf{88.84} \\
TCIL-Lite                                               & 14.5     & 77.50 & \multicolumn{1}{c|}{67.30} & 93.60  & 87.94                       \\ \bottomrule
\end{tabular}
}
\caption{Top-1 and top-5 accuracy comparison on ImageNet100-B0 in rehearsal setting.}
\label{IM100-B0}
\end{table}

\subsection{Results and Discussion}
\textbf{Rehearsal Setting.} 
The results on CIFAR100 and ImageNet100 with different protocols are given in Tab.\ref{Cifar-B0} and Tab.\ref{IM100-B0}, respectively. In both CIFAR100-B0 and CIFAR100-B50, DEA-based models (e.g., DER, DyTox and TCIL) outperform non-DEA-based models by large margins at all task splits. They also exhibit smaller accuracy decline with the accumulation of incremental steps, showing the effectiveness of DEA-based methods in reducing the forgetting of old knowledge. When looking into DEA-based models, TCIL gains certain accuracy improvements but with increased model size, especially compared with DyTox. However, TCIL-Lite largely alleviates this dilemma. Although slightly inferior to TCIL at small steps (i.e., 2 or 5 steps), TCIL-Lite gradually approaches and even outperforms TCIL with the increase of incremental steps, indicating that the pruning almost does not affect the network capability in rehearsal setting. Moreover, when compared with DER, the previous leading method in the DEA family, TCIL-Lite consistently has better accuracy and fewer parameters. Note that similar comparison results are also observed in ImageNet100-B0, TCIL and TCIL-Lite suppress existing models in most cases. In Fig.\ref{img2}, four line charts with different configurations are depicted to show the detailed accuracy evolution with the incremental steps (More line charts are given in the supplement). TCIL and TCIL-Lite are always with the slowest forgetting rate and ranking top-tier. The results clearly demonstrate the superiority of TCIL in rehearsal setting. 

%Moreover, as opposed to non-DEA models that have constant model size, parameters of DER and TCIL grow with incremental steps, which is a drawback in real-world applications. As a solution, TCIL-Lite, the lite version TCIL, is obtained by applying parameter pruning techniques described in~\cite{e12}. We set TCIL-Lite nearly four times smaller than TCIL such that the model with 10 incremental steps has parameters similar to non-DEA-based models. 

\begin{table}
\centering

\resizebox{1.0\linewidth}{!}{
\begin{tabular}{@{}l|cccccc|ccc@{}}
\toprule
\multicolumn{1}{c|}{\multirow{3}{*}{\textbf{Methods}}} & \multicolumn{6}{c|}{CIFAR100 B0}                                                                                             & \multicolumn{3}{c}{ImageNet100-B0}                 \\ \cmidrule(l){2-10} 
\multicolumn{1}{c|}{}                                  & \multicolumn{3}{c|}{5 steps}                                            & \multicolumn{3}{c|}{10 steps}                      & \multicolumn{3}{c}{10 steps}                       \\ \cmidrule(l){2-10} 
\multicolumn{1}{c|}{}                                  & \textbf{\#Paras} & \textbf{Avg}   & \multicolumn{1}{c|}{\textbf{Last}}  & \textbf{\#Paras} & \textbf{Avg}   & \textbf{Last}  & \textbf{\#Paras} & \textbf{Avg}   & \textbf{Last}  \\ \midrule
Finetuning                                             & 11.2             & 32.89          & \multicolumn{1}{c|}{11.42}          & 11.2             & 20.71          & 6.36           & 11.2             & 28.48          & 9.48           \\
LwF-MC~\shortcite{t54}           & -                & 54.74          & \multicolumn{1}{c|}{35.42}          & -                & 45.08          & 24.42          & -                & 65.45          & 37.42          \\
BiC~\shortcite{t69}              & 11.2             & 39.31          & \multicolumn{1}{c|}{16.70}          & 11.2             & 25.47          & 8.57           & 11.2             & 36.24          & 10.02          \\
LwM~\shortcite{p24}              & -                & 63.67          & \multicolumn{1}{c|}{49.36}          & -                & 45.68          & 26.14          & -                & -              & -              \\
WA~\shortcite{d39}               & 11.2             & 59.14          & \multicolumn{1}{c|}{30.61}          & 11.2             & 37.58          & 5.53           & 11.2             & 65.52          & 21.18          \\
DER~\shortcite{t70}              & 33.6             & 39.93          & \multicolumn{1}{c|}{17.31}          & 61.6             & 26.53          & 8.88           & 61.6             & 29.15          & 9.96           \\ \midrule
TCIL                                                   & 34.3             & \textbf{63.92} & \multicolumn{1}{c|}{\textbf{53.08}} & 64.1             & \textbf{56.57} & \textbf{39.91} & 64.1             & \textbf{84.05} & \textbf{76.82} \\
TCIL-Lite                                              & 8.3              & 63.72          & \multicolumn{1}{c|}{51.82}          & 14.5             & 56.18          & 38.34          & 14.5             & 82.81          & 69.80          \\ \bottomrule
\end{tabular}
}
\caption{Accuracy comparison on CIFAR100 and ImageNet100 in non-rehearsal setting. LwM and LwF-MC results come from~\cite{p24}. Other results are reported based on open source codes or our implementation.}
\label{mem_free}
\end{table}

\begin{table*}[htb]
    \centering
    \resizebox{\linewidth}{!}{
\begin{tabular}{@{}cc|ccccccccc|ccccccccc|ccc@{}}
\toprule
\multicolumn{2}{c|}{}                                   & \multicolumn{9}{c|}{\textbf{CIFAR100-B0}}                                                                                                                                                                       & \multicolumn{9}{c|}{\textbf{CIFAR100-B50}}                                                                                                                                                       & \multicolumn{3}{c}{\textbf{ImageNet100-B0}}      \\ \cmidrule(l){3-23} 
\multicolumn{2}{c|}{}                                   & \multicolumn{3}{c|}{5   steps}                                        & \multicolumn{3}{c|}{10 steps}                                                        & \multicolumn{3}{c|}{20 steps}                    & \multicolumn{3}{c|}{2   steps}                                        & \multicolumn{3}{c|}{5 steps}                                          & \multicolumn{3}{c|}{10 steps}                    & \multicolumn{3}{c}{10 steps}                     \\ \cmidrule(l){3-23} 
\multicolumn{2}{c|}{\multirow{-3}{*}{\textbf{Methods}}} & 2000           & 1000           & \multicolumn{1}{c|}{0}              & 2000           & 1000                          & \multicolumn{1}{c|}{0}              & 2000           & 1000           & 0              & 2000           & 1000           & \multicolumn{1}{c|}{0}              & 2000           & 1000           & \multicolumn{1}{c|}{0}              & 2000           & 1000           & 0              & 2000           & 1000           & 0              \\ \midrule
\multicolumn{1}{c|}{}                          & \textbf{Avg}    & 76.80          & 73.67          & \multicolumn{1}{c|}{39.93}          & 75.36          & 71.10                         & \multicolumn{1}{c|}{26.53}          & 74.09          & 70.34          & 16.96          & 74.61          & 72.50          & \multicolumn{1}{c|}{43.89}          & 73.21          & 70.17          & \multicolumn{1}{c|}{23.31}          & 72.81          & 69.67          & 13.11          & 93.23          & 78.06          & 29.15          \\
\multicolumn{1}{c|}{\multirow{-2}{*}{DER~\shortcite{t70}}}     & \textbf{Last}   & 68.32          & 62.22          & \multicolumn{1}{c|}{17.31}          & 65.22          & 56.77 & \multicolumn{1}{c|}{8.88}           & 62.48          & 53.32          & 4.80            & 68.84          & 64.70           & \multicolumn{1}{c|}{21.45}          & 73.21          & 59.99          & \multicolumn{1}{c|}{8.85}           & 65.45          & 56.98          & 4.80            & 87.52          & 53.96          & 9.96           \\ \midrule
\multicolumn{1}{c|}{}                          & \textbf{Avg}    & \textbf{77.72} & \textbf{75.64} & \multicolumn{1}{c|}{\textbf{63.92}} & \textbf{77.30} & \textbf{74.13}                & \multicolumn{1}{c|}{\textbf{56.57}} & \textbf{75.11} & \textbf{71.85} & \textbf{46.74} & \textbf{76.42} & \textbf{75.17} & \multicolumn{1}{c|}{\textbf{65.78}} & \textbf{74.88} & \textbf{74.34} & \multicolumn{1}{c|}{\textbf{63.18}} & \textbf{73.72} & \textbf{71.91} & \textbf{47.52} & \textbf{94.17} & \textbf{93.66} & \textbf{84.05} \\
\multicolumn{1}{c|}{\multirow{-2}{*}{TCIL}}    & \textbf{Last}   & \textbf{69.58} & \textbf{67.00}    & \multicolumn{1}{c|}{\textbf{53.08}} & \textbf{66.41} & \textbf{62.87}                & \multicolumn{1}{c|}{\textbf{39.91}} & \textbf{63.54} & \textbf{59.07} & \textbf{27.11} & \textbf{71.91} & \textbf{70.38} & \multicolumn{1}{c|}{\textbf{54.85}} & \textbf{74.88} & \textbf{67.90}  & \multicolumn{1}{c|}{\textbf{45.39}} & \textbf{66.36} & \textbf{64.79} & \textbf{255.4} & \textbf{88.84} & \textbf{86.04} & \textbf{76.82} \\ \bottomrule
\end{tabular}
}
\caption{Accuracy comparison with different memory budget (number of exemplars). We report the top-1 accuracy on CIFAR100 and the top-5 accuracy on ImageNet100.}
\label{mem1}
\end{table*}

\noindent\textbf{Non-Rehearsal Setting.}
Table~\ref{mem_free} shows the results on CIFAR100-B0 and ImageNet100-B0 with different incremental steps. The results are generally worse than those with the rehearsal memory and experience even sharp decreases as the increase of incremental steps, showing the positive effect of reserving certain old data. When comparing existing models, TCIL surpasses LwM, the previous state of the art, by 3.72\% and 13.7\% at "Last" on CIFAR100-B0 with 5 and 10 steps, respectively. Moreover, TCIL outperforms DER by significant larger margins compared to the rehearsal case. It implies that TCIL is less sensitive to the availability of rehearsal memory, which is an advantage in many applications. Compared TCIL with TCIL-Lite, TCIL is more robust as the step accumulated. It can be explained as the generalization ability of TCIL-Lite is worse in challenging cases due to the limited network capacity. In non-rehearsal scenario, TCIL only distills from the outputted logits, where much less knowledge could be dug. Nevertheless, TCIL-Lite still performs significantly better than DER in all the listed protocols while with few parameters. The observations reveal advantages of the TCIL family from another angle and again demonstrate their effectiveness.

 \begin{table}
    \centering
    \resizebox{.7\columnwidth}{!}{
    \begin{tabular}{@{}cccccc@{}}
    \toprule
    \textbf{DEA} & \textbf{Div Loss} & \textbf{CR} & \textbf{MLKD} & \textbf{FFM} & \textbf{Avg} \\ \midrule
     &    &  &   &  & 61.84        \\
     \Checkmark   & & & &  & 69.45      \\
     \Checkmark  & \Checkmark  & &  &   & 70.59      \\
    \Checkmark  & \Checkmark  & \Checkmark & &  & 73.43      \\
    \Checkmark                              & \Checkmark                      & \Checkmark                      & \Checkmark                               &  & 75.80      \\
    \Checkmark                              & \Checkmark                      & \Checkmark                      & \Checkmark                               & \Checkmark                        & 77.30      \\ \bottomrule
    \end{tabular}
    }
    \caption{Ablations on the components of TCIL}
    \label{Ablations}
\end{table}

\subsection{Ablation Study and Error Analysis}
We ablate the components in TCIL to validate their utilities and Tab.~\ref{Ablations} give the result, where ResNet-18 with 10 incremental steps and the rehearsal strategy is taken as the baseline. First, equipping the DEA generates 7.61\% accuracy gains. Then, applying the divergence loss to encourage feature extractors to better distinguish between old-new classes gives marginal improvements. In the following, applying CR and MLKD further gives 2.84\% and 2.37\% accuracy improvements, indicating that the two upgrades are good at suppressing task confusion in the DEA framework. Finally, FFM also reports an improvement of 1.5\% by injecting the combined attention mechanism. The ablation basically verifies the effectiveness of the employed components.

%\textbf{ERROR Analysis}
To analyze the task confusion within DEAs, we perform experiments on CIFAR100-B0 with 10 incremental steps and group the errors into three types, i.e., ITC, ONC and within-task confusion (WTC). ResNet-18 with DEA and rehearsal strategy are employed as the baseline, i.e., the second line in Tab.~\ref{Ablations}. While CR and TCIL are compared, which correspond to the fourth and sixth lines in Tab.~\ref{Ablations}, respectively. Task confusion in the form of error statistics are summarized in Tab.~\ref{ERROR Analysis}. As can be seen, the three methods gradually exhibit differences with the increase of steps. ONC errors have been reduced significantly by applying CR. However, it also leads to an increase of ITC errors. We argue that these newly added ITC errors are samples that are prone to ONC, and were previously classified as ONC errors due to severe task-tendency bias in the baseline. Nevertheless, compared to CR, TCIL further reduces all three kinds of errors. These results intuitively show that TCIL can effectively reduce task confusion by mitigating ONC and ITC in a targeted manner. On the other hand, WTC errors account for a small proportion of the total number of errors, less than 10\% at the last step. This result indicates that ITC and ONC are the main cause of catastrophic forgetting, which is different from the case of fixed network structure that often suffers from weight drift~\cite{e2, t38}. Our exploration basically verifies that to address the catastrophic forgetting in DEAs, more attention needs to be paid to the task confusion,  especially ITC and ONC.

 \begin{table}
    \centering
    \resizebox{\linewidth}{!}{
\begin{tabular}{cc|cccccccccc}
\hline
\multirow{2}{*}{\textbf{Methods}} & \multirow{2}{*}{\textbf{Error}} & \multicolumn{10}{c}{\textbf{Incremental tasks}}               \\ \cline{3-12} 
                                  &                                 & 1  & 2   & 3   & 4   & 5   & 6    & 7    & 8    & 9    & 10   \\ \hline
\multirow{3}{*}{baseline}         & WTC                             & 75 & 133 & 169 & 164 & 191 & 169  & 226  & 182  & 207  & 254  \\
                                  & ONC                             & 0  & 162 & 450 & 750 & 856 & 1034 & 1597 & 1834 & 2053 & 2943 \\
                                  & ITC                             & 0  & 0   & 67  & 237 & 524 & 794  & 824  & 1200 & 1481 & 1538 \\ \hline
\multirow{3}{*}{CR}               & WTC                             & 75 & 132 & 173 & 169 & 199 & 194  & 232  & 228  & 248  & 271  \\
                                  & ONC                             & 0  & 170 & 390 & 583 & 541 & 515  & 739  & 743  & 647  & 832  \\
                                  & ITC                             & 0  & 0   & 75  & 291 & 670 & 1023 & 1249 & 1733 & 2254 & 2718 \\ \hline
\multirow{3}{*}{TCIL}              & WTC                             & 75 & 120 & 157 & 177 & 212 & 211  & 223  & 221  & 238  & 248  \\
                                  & ONC                             & 0  & 141 & 284 & 382 & 309 & 345  & 538  & 476  & 453  & 709  \\
                                  & ITC                             & 0  & 0   & 92  & 309 & 630 & 879  & 1156 & 1619 & 2046 & 2391 \\ \hline
\end{tabular}
    }
\caption{Error statistics on CIFAR100-B0. ITC, ONC, WTC denote inter-task confusion, old-new confusion, within-task confusion, respectively.},
\label{ERROR Analysis}
\end{table}

We set up a controlled trial with DER to illustrate TCIL is much less dependent on rehearsal. It is clearly shown in Tab.~\ref{mem1} that as the memory size decreases, the gap between TCIL and DER in accuracy becomes larger no matter the dataset, evaluation protocols and task splits. The result strongly demonstrates that TCIL proposed in this paper can alleviate the problem that DER relies much on the rehearsal mechanism under the structure of dynamic extension.

\section{Conclusion}
We have inspected class incremental learning from the task confusion angle, where ITC and ONC have been pointed out as two major causes of catastrophic forgetting in DEAs. As a consequence, TCIL is presented. It develops a multi-level knowledge distillation to promote knowledge propagation and utilization, while attention mechanism and classifier re-scoring strategy are also taken into account. The experiments conducted on CIFAR100 and ImageNet100 basically verify our proposal. TCIL and TCIL-lite consistently report state-of-the-art accuracy. They are also more robust with the accumulation of incremental steps and less sensitive to the availability of rehearsal memory. The error statistics imply that task confusion is largely reduced by TCIL especially ONC errors. The strength of error reduction, though, is still limited by the fact that the knowledge in different sub-networks is still not being enough shared and fused. Thus, future work includes the exploration of more effective knowledge sharing and utilization protocols. Meanwhile, we are also interested in extending TCIL to other research scenarios, e.g., the label sets are not exclusive.

\section{Acknowledgments}
This project was supported by National Key R\&D Program of China (No. 2021ZD0112804) and in part by the National Natural Science Foundation of China (No. 62172103,  62102092)

\bibliography{aaai23}

\begin{thebibliography}{53}
\providecommand{\natexlab}[1]{#1}

\bibitem[{Aljundi et~al.(2018)Aljundi, Babiloni, Elhoseiny, Rohrbach, and
  Tuytelaars}]{e2}
Aljundi, R.; Babiloni, F.; Elhoseiny, M.; Rohrbach, M.; and Tuytelaars, T.
  2018.
\newblock Memory aware synapses: Learning what (not) to forget.
\newblock In \emph{Proceedings of the European Conference on Computer Vision},
  139--154.

\bibitem[{Belouadah and Popescu(2019)}]{v25}
Belouadah, E.; and Popescu, A. 2019.
\newblock Il2m: Class incremental learning with dual memory.
\newblock In \emph{Proceedings of the IEEE/CVF International Conference on
  Computer Vision}, 583--592.

\bibitem[{Castro et~al.(2018)Castro, Mar{\'\i}n-Jim{\'e}nez, Guil, Schmid, and
  Alahari}]{w5}
Castro, F.~M.; Mar{\'\i}n-Jim{\'e}nez, M.~J.; Guil, N.; Schmid, C.; and
  Alahari, K. 2018.
\newblock End-to-end incremental learning.
\newblock In \emph{Proceedings of the European conference on computer vision},
  233--248.

\bibitem[{Chen et~al.(2021)Chen, Liu, Zhao, and Jia}]{e14}
Chen, P.; Liu, S.; Zhao, H.; and Jia, J. 2021.
\newblock Distilling knowledge via knowledge review.
\newblock In \emph{Proceedings of the IEEE/CVF Conference on Computer Vision
  and Pattern Recognition}, 5008--5017.

\bibitem[{Collier et~al.(2020)Collier, Kokiopoulou, Gesmundo, and Berent}]{t11}
Collier, M.; Kokiopoulou, E.; Gesmundo, A.; and Berent, J. 2020.
\newblock Routing networks with co-training for continual learning.
\newblock \emph{arXiv preprint arXiv:2009.04381}.

\bibitem[{De~Lange et~al.(2019)De~Lange, Aljundi, Masana, Parisot, Jia,
  Leonardis, Slabaugh, and Tuytelaars}]{s14}
De~Lange, M.; Aljundi, R.; Masana, M.; Parisot, S.; Jia, X.; Leonardis, A.;
  Slabaugh, G.; and Tuytelaars, T. 2019.
\newblock Continual learning: A comparative study on how to defy forgetting in
  classification tasks.
\newblock \emph{arXiv preprint arXiv:1909.08383}, 2(6).

\bibitem[{Dhar et~al.(2019)Dhar, Singh, Peng, Wu, and Chellappa}]{p24}
Dhar, P.; Singh, R.~V.; Peng, K.-C.; Wu, Z.; and Chellappa, R. 2019.
\newblock Learning without memorizing.
\newblock In \emph{Proceedings of the IEEE/CVF Conference on Computer Vision
  and Pattern Recognition}, 5138--5146.

\bibitem[{Douillard et~al.(2020)Douillard, Cord, Ollion, Robert, and
  Valle}]{t19}
Douillard, A.; Cord, M.; Ollion, C.; Robert, T.; and Valle, E. 2020.
\newblock Podnet: Pooled outputs distillation for small-tasks incremental
  learning.
\newblock In \emph{European Conference on Computer Vision}, 86--102. Springer.

\bibitem[{Douillard et~al.(2022)Douillard, Ram{\'e}, Couairon, and Cord}]{t0}
Douillard, A.; Ram{\'e}, A.; Couairon, G.; and Cord, M. 2022.
\newblock Dytox: Transformers for continual learning with dynamic token
  expansion.
\newblock In \emph{Proceedings of the IEEE/CVF Conference on Computer Vision
  and Pattern Recognition}, 9285--9295.

\bibitem[{Fernando et~al.(2017)Fernando, Banarse, Blundell, Zwols, Ha, Rusu,
  Pritzel, and Wierstra}]{t23}
Fernando, C.; Banarse, D.; Blundell, C.; Zwols, Y.; Ha, D.; Rusu, A.~A.;
  Pritzel, A.; and Wierstra, D. 2017.
\newblock Pathnet: Evolution channels gradient descent in super neural
  networks.
\newblock \emph{arXiv preprint arXiv:1701.08734}.

\bibitem[{Fletcher, Venkatasubramanian, and Joshi(2008)}]{e15}
Fletcher, P.~T.; Venkatasubramanian, S.; and Joshi, S. 2008.
\newblock Robust statistics on Riemannian manifolds via the geometric median.
\newblock In \emph{IEEE Conference on Computer Vision and Pattern Recognition},
  1--8. IEEE.

\bibitem[{Golkar, Kagan, and Cho(2019)}]{t27}
Golkar, S.; Kagan, M.; and Cho, K. 2019.
\newblock Continual learning via neural pruning.
\newblock \emph{arXiv preprint arXiv:1903.04476}.

\bibitem[{Goodfellow et~al.(2013)Goodfellow, Mirza, Xiao, Courville, and
  Bengio}]{m12}
Goodfellow, I.~J.; Mirza, M.; Xiao, D.; Courville, A.; and Bengio, Y. 2013.
\newblock An Empirical Investigation of Catastrophic Forgetting in
  Gradient-Based Neural Networks.
\newblock \emph{Computer Science}, 84(12): 1387--91.

\bibitem[{He et~al.(2019)He, Liu, Wang, Hu, and Yang}]{e12}
He, Y.; Liu, P.; Wang, Z.; Hu, Z.; and Yang, Y. 2019.
\newblock Filter pruning via geometric median for deep convolutional neural
  networks acceleration.
\newblock In \emph{Proceedings of the IEEE/CVF Conference on Computer Vision
  and Pattern Recognition}, 4340--4349.

\bibitem[{Hinton, Vinyals, and Dean(2015)}]{r9}
Hinton, G.; Vinyals, O.; and Dean, J. 2015.
\newblock Distilling the Knowledge in a Neural Network.
\newblock \emph{Computer Science}, 14(7): 38--39.

\bibitem[{Hou et~al.(2019)Hou, Pan, Loy, Wang, and Lin}]{t32}
Hou, S.; Pan, X.; Loy, C.~C.; Wang, Z.; and Lin, D. 2019.
\newblock Learning a unified classifier incrementally via rebalancing.
\newblock In \emph{Proceedings of the IEEE/CVF Conference on Computer Vision
  and Pattern Recognition}, 831--839.

\bibitem[{Hu, Shen, and Sun(2018)}]{c28}
Hu, J.; Shen, L.; and Sun, G. 2018.
\newblock Squeeze-and-excitation networks.
\newblock In \emph{Proceedings of the IEEE conference on computer vision and
  pattern recognition}, 7132--7141.

\bibitem[{Hung et~al.(2019)Hung, Tu, Wu, Chen, Chan, and Chen}]{t33}
Hung, C.-Y.; Tu, C.-H.; Wu, C.-E.; Chen, C.-H.; Chan, Y.-M.; and Chen, C.-S.
  2019.
\newblock Compacting, picking and growing for unforgetting continual learning.
\newblock \emph{Advances in Neural Information Processing Systems}, 32.

\bibitem[{Kirkpatrick et~al.(2017)Kirkpatrick, Pascanu, Rabinowitz, Veness,
  Desjardins, Rusu, Milan, Quan, Ramalho, Grabska-Barwinska et~al.}]{t38}
Kirkpatrick, J.; Pascanu, R.; Rabinowitz, N.; Veness, J.; Desjardins, G.; Rusu,
  A.~A.; Milan, K.; Quan, J.; Ramalho, T.; Grabska-Barwinska, A.; et~al. 2017.
\newblock Overcoming catastrophic forgetting in neural networks.
\newblock \emph{Proceedings of the national academy of sciences}, 114(13):
  3521--3526.

\bibitem[{Krizhevsky(2009)}]{e6}
Krizhevsky, A. 2009.
\newblock Learning Multiple Layers of Features from Tiny Images.
\newblock \emph{Master's thesis, University of Tront}.

\bibitem[{Lange et~al.(2020)Lange, Jia, Parisot, Leonardis, Slabaugh, and
  Tuytelaars}]{e0}
Lange, M.~D.; Jia, X.; Parisot, S.; Leonardis, A.; Slabaugh, G.; and
  Tuytelaars, T. 2020.
\newblock Unsupervised model personalization while preserving privacy and
  scalability: An open problem.
\newblock In \emph{Proceedings of the IEEE/CVF Conference on Computer Vision
  and Pattern Recognition}, 14463--14472.

\bibitem[{Li et~al.(2017)Li, Jun, Fei, and Li}]{d19}
Li, L.; Jun, Z.; Fei, J.; and Li, S. 2017.
\newblock An incremental face recognition system based on deep learning.
\newblock In \emph{2017 Fifteenth IAPR International Conference on Machine
  Vision Applications (MVA)}, 238--241.

\bibitem[{Li et~al.(2019)Li, Wang, Hu, and Yang}]{e10}
Li, X.; Wang, W.; Hu, X.; and Yang, J. 2019.
\newblock Selective kernel networks.
\newblock In \emph{Proceedings of the IEEE/CVF Conference on Computer Vision
  and Pattern Recognition}, 510--519.

\bibitem[{Li and Hoiem(2017)}]{w20}
Li, Z.; and Hoiem, D. 2017.
\newblock Learning without forgetting.
\newblock \emph{IEEE transactions on pattern analysis and machine
  intelligence}, 40(12): 2935--2947.

\bibitem[{Li et~al.(2021)Li, Zhong, Liu, Wang, and Zheng}]{t47}
Li, Z.; Zhong, C.; Liu, S.; Wang, R.; and Zheng, W.-S. 2021.
\newblock Preserving earlier knowledge in continual learning with the help of
  all previous feature extractors.
\newblock \emph{arXiv preprint arXiv:2104.13614}.

\bibitem[{Lu et~al.(2022)Lu, Xie, Liu, and Zhang}]{lu2022bridging}
Lu, Z.; Xie, H.; Liu, C.; and Zhang, Y. 2022.
\newblock Bridging the Gap Between Vision Transformers and Convolutional Neural
  Networks on Small Datasets.
\newblock In \emph{Advances in Neural Information Processing Systems}.

\bibitem[{Masana et~al.(2020)Masana, Liu, Twardowski, Menta, Bagdanov, and
  van~de Weijer}]{v0}
Masana, M.; Liu, X.; Twardowski, B.; Menta, M.; Bagdanov, A.~D.; and van~de
  Weijer, J. 2020.
\newblock Class-incremental learning: survey and performance evaluation on
  image classification.
\newblock \emph{arXiv preprint arXiv:2010.15277}.

\bibitem[{Min et~al.(2020)Min, Yao, Xie, Zha, and Zhang}]{min2020multi}
Min, S.; Yao, H.; Xie, H.; Zha, Z.-J.; and Zhang, Y. 2020.
\newblock Multi-objective matrix normalization for fine-grained visual
  recognition.
\newblock \emph{IEEE Transactions on Image Processing}, 29: 4996--5009.

\bibitem[{Mittal, Galesso, and Brox(2021)}]{e13}
Mittal, S.; Galesso, S.; and Brox, T. 2021.
\newblock Essentials for class incremental learning.
\newblock In \emph{Proceedings of the IEEE/CVF Conference on Computer Vision
  and Pattern Recognition}, 3513--3522.

\bibitem[{Parisi et~al.(2019)Parisi, Kemker, Part, Kanan, and Wermter}]{s43}
Parisi, G.~I.; Kemker, R.; Part, J.~L.; Kanan, C.; and Wermter, S. 2019.
\newblock Continual lifelong learning with neural networks: A review.
\newblock \emph{Neural Networks}, 113: 54--71.

\bibitem[{Passalis, Tzelepi, and Tefas(2020)}]{r23}
Passalis, N.; Tzelepi, M.; and Tefas, A. 2020.
\newblock Probabilistic knowledge transfer for lightweight deep representation
  learning.
\newblock \emph{IEEE Transactions on Neural Networks and Learning Systems},
  32(5): 2030--2039.

\bibitem[{Pierre(2018)}]{d25}
Pierre, J.~M. 2018.
\newblock Incremental lifelong deep learning for autonomous vehicles.
\newblock In \emph{2018 21st international conference on intelligent
  transportation systems}, 3949--3954. IEEE.

\bibitem[{Rajasegaran et~al.(2019)Rajasegaran, Hayat, Khan, Khan, Shao, and
  Yang}]{t52}
Rajasegaran, J.; Hayat, M.; Khan, S.; Khan, F.~S.; Shao, L.; and Yang, M.-H.
  2019.
\newblock An adaptive random path selection approach for incremental learning.
\newblock \emph{arXiv preprint arXiv:1906.01120}.

\bibitem[{Rebuffi et~al.(2017)Rebuffi, Kolesnikov, Sperl, and Lampert}]{t54}
Rebuffi, S.-A.; Kolesnikov, A.; Sperl, G.; and Lampert, C.~H. 2017.
\newblock icarl: Incremental classifier and representation learning.
\newblock In \emph{Proceedings of the IEEE conference on Computer Vision and
  Pattern Recognition}, 2001--2010.

\bibitem[{Russakovsky et~al.(2015)Russakovsky, Deng, Su, Krause, Satheesh, Ma,
  Huang, Karpathy, Khosla, Bernstein et~al.}]{e7}
Russakovsky, O.; Deng, J.; Su, H.; Krause, J.; Satheesh, S.; Ma, S.; Huang, Z.;
  Karpathy, A.; Khosla, A.; Bernstein, M.; et~al. 2015.
\newblock Imagenet large scale visual recognition challenge.
\newblock \emph{International journal of computer vision}, 115(3): 211--252.

\bibitem[{Rusu et~al.(2016)Rusu, Rabinowitz, Desjardins, Soyer, Kirkpatrick,
  Kavukcuoglu, Pascanu, and Hadsell}]{t56}
Rusu, A.~A.; Rabinowitz, N.~C.; Desjardins, G.; Soyer, H.; Kirkpatrick, J.;
  Kavukcuoglu, K.; Pascanu, R.; and Hadsell, R. 2016.
\newblock Progressive neural networks.
\newblock \emph{arXiv preprint arXiv:1606.04671}.

\bibitem[{Shin et~al.(2017)Shin, Lee, Kim, and Kim}]{v29}
Shin, H.; Lee, J.~K.; Kim, J.; and Kim, J. 2017.
\newblock Continual learning with deep generative replay.
\newblock In \emph{Advances in neural information processing systems},
  2994–3003.

\bibitem[{Thrun(1995)}]{m32}
Thrun, S. 1995.
\newblock Is learning the n-th thing any easier than learning the first?
\newblock \emph{Advances in neural information processing systems}, 8.

\bibitem[{Wang et~al.(2018)Wang, Fan, Wang, Jiao, and Schiele}]{e11}
Wang, H.; Fan, Y.; Wang, Z.; Jiao, L.; and Schiele, B. 2018.
\newblock Parameter-free spatial attention network for person
  re-identification.
\newblock \emph{arXiv preprint arXiv:1811.12150}.

\bibitem[{Wen, Tran, and Ba(2020)}]{t67}
Wen, Y.; Tran, D.; and Ba, J. 2020.
\newblock Batchensemble: an alternative approach to efficient ensemble and
  lifelong learning.
\newblock \emph{arXiv preprint arXiv:2002.06715}.

\bibitem[{Woo et~al.(2018)Woo, Park, Lee, and Kweon}]{c0}
Woo, S.; Park, J.; Lee, J.-Y.; and Kweon, I.~S. 2018.
\newblock Cbam: Convolutional block attention module.
\newblock In \emph{Proceedings of the European conference on computer vision},
  3--19.

\bibitem[{Wu et~al.(2018)Wu, Herranz, Liu, van~de Weijer, Raducanu et~al.}]{e3}
Wu, C.; Herranz, L.; Liu, X.; van~de Weijer, J.; Raducanu, B.; et~al. 2018.
\newblock Memory replay gans: Learning to generate new categories without
  forgetting.
\newblock \emph{Advances in Neural Information Processing Systems}, 31.

\bibitem[{Wu et~al.(2019)Wu, Chen, Wang, Ye, Liu, Guo, and Fu}]{t69}
Wu, Y.; Chen, Y.; Wang, L.; Ye, Y.; Liu, Z.; Guo, Y.; and Fu, Y. 2019.
\newblock Large scale incremental learning.
\newblock In \emph{Proceedings of the IEEE/CVF Conference on Computer Vision
  and Pattern Recognition}, 374--382.

\bibitem[{Xiang et~al.(2019)Xiang, Fu, Ji, and Huang}]{e5}
Xiang, Y.; Fu, Y.; Ji, P.; and Huang, H. 2019.
\newblock Incremental learning using conditional adversarial networks.
\newblock In \emph{Proceedings of the IEEE/CVF International Conference on
  Computer Vision}, 6619--6628.

\bibitem[{Yan, Xie, and He(2021)}]{t70}
Yan, S.; Xie, J.; and He, X. 2021.
\newblock Der: Dynamically expandable representation for class incremental
  learning.
\newblock In \emph{Proceedings of the IEEE/CVF Conference on Computer Vision
  and Pattern Recognition}, 3014--3023.

\bibitem[{Yang et~al.(2022{\natexlab{a}})Yang, Zhou, Shi, Wu, and
  Wang}]{yang2022rd}
Yang, D.; Zhou, Y.; Shi, W.; Wu, D.; and Wang, W. 2022{\natexlab{a}}.
\newblock RD-IOD: Two-Level Residual-Distillation-Based Triple-Network for
  Incremental Object Detection.
\newblock \emph{ACM Transactions on Multimedia Computing, Communications, and
  Applications (TOMM)}, 18(1): 1--23.

\bibitem[{Yang et~al.(2022{\natexlab{b}})Yang, Zhou, Zhang, Sun, Wu, Wang, and
  Ye}]{yang2022multi}
Yang, D.; Zhou, Y.; Zhang, A.; Sun, X.; Wu, D.; Wang, W.; and Ye, Q.
  2022{\natexlab{b}}.
\newblock Multi-view correlation distillation for incremental object detection.
\newblock \emph{Pattern Recognition}, 131: 108863.

\bibitem[{Yu et~al.(2020)Yu, Twardowski, Liu, Herranz, Wang, Cheng, Jui, and
  Weijer}]{e4}
Yu, L.; Twardowski, B.; Liu, X.; Herranz, L.; Wang, K.; Cheng, Y.; Jui, S.; and
  Weijer, J. v.~d. 2020.
\newblock Semantic drift compensation for class-incremental learning.
\newblock In \emph{Proceedings of the IEEE/CVF Conference on Computer Vision
  and Pattern Recognition}, 6982--6991.

\bibitem[{Zenke, Poole, and Ganguli(2017)}]{e1}
Zenke, F.; Poole, B.; and Ganguli, S. 2017.
\newblock Continual learning through synaptic intelligence.
\newblock In \emph{International Conference on Machine Learning}, 3987--3995.
  PMLR.

\bibitem[{Zhao et~al.(2020)Zhao, Xiao, Gan, Zhang, and Xia}]{d39}
Zhao, B.; Xiao, X.; Gan, G.; Zhang, B.; and Xia, S.-T. 2020.
\newblock Maintaining discrimination and fairness in class incremental
  learning.
\newblock In \emph{Proceedings of the IEEE/CVF Conference on Computer Vision
  and Pattern Recognition}, 13208--13217.

\bibitem[{Zhou et~al.(2019)Zhou, Mai, Zhang, Xu, Wu, and Davis}]{w39}
Zhou, P.; Mai, L.; Zhang, J.; Xu, N.; Wu, Z.; and Davis, L.~S. 2019.
\newblock M2kd: Multi-model and multi-level knowledge distillation for
  incremental learning.
\newblock \emph{arXiv preprint arXiv:1904.01769}.

\bibitem[{Zhu et~al.(2021)Zhu, Zhang, Wang, Yin, and Liu}]{zhu2021prototype}
Zhu, F.; Zhang, X.-Y.; Wang, C.; Yin, F.; and Liu, C.-L. 2021.
\newblock Prototype augmentation and self-supervision for incremental learning.
\newblock In \emph{Proceedings of the IEEE/CVF Conference on Computer Vision
  and Pattern Recognition}, 5871--5880.

\bibitem[{Zhu et~al.(2022)Zhu, Zhai, Cao, Luo, and Zha}]{zhu2022self}
Zhu, K.; Zhai, W.; Cao, Y.; Luo, J.; and Zha, Z.-J. 2022.
\newblock Self-Sustaining Representation Expansion for Non-Exemplar
  Class-Incremental Learning.
\newblock In \emph{Proceedings of the IEEE/CVF Conference on Computer Vision
  and Pattern Recognition}, 9296--9305.

\end{thebibliography}

\end{document}